\documentclass{sig-alternate-2013}

\newfont{\mycrnotice}{ptmr8t at 7pt}
\newfont{\myconfname}{ptmri8t at 7pt}
\let\crnotice\mycrnotice%
\let\confname\myconfname%

\permission{Permission to make digital or hard copies of all or part of this work for personal or classroom use is granted without fee provided that copies are not made or distributed for profit or commercial advantage and that copies bear this notice and the full citation on the first page. Copyrights for components of this work owned by others than ACM must be honored. Abstracting with credit is permitted. To copy otherwise, or republish, to post on servers or to redistribute to lists, requires prior specific permission and/or a fee. Request permissions from Permissions@acm.org.}
\conferenceinfo{KDD'15,}{August 10-13, 2015, Sydney, NSW, Australia.}
\copyrightetc{\copyright~2015 ACM. ISBN \the\acmcopyr}
\crdata{978-1-4503-3664-2/15/08\ ...\$15.00.\\
DOI: http://dx.doi.org/10.1145/2783258.2788616.}

\usepackage{framed}
\usepackage{multirow}
\usepackage{graphics}
\usepackage{url}
\usepackage{wrapfig}
\usepackage{algorithmic} 
\usepackage{algorithm} 
\usepackage{float} 
\usepackage{amsmath}
\usepackage{amssymb}
\usepackage{amsfonts}

\usepackage[table]{xcolor}
\usepackage{epsfig}
\usepackage{booktabs}

\definecolor{lightgray}{gray}{0.85}

\numberwithin{equation}{section}

\newcommand{\R}{\mathbb{R}}

\usepackage{subfig, float, epsfig, epstopdf, balance, algorithm}
\usepackage{helvet}
\usepackage{courier}
\usepackage{multirow}
\usepackage{mathtools}
\usepackage{hyperref}

\def\P{\mathbb{P}}
\def\T{\mathrm{T}}

\begin{document}

\title{Gender and Interest Targeting for \\Sponsored Post Advertising at Tumblr}

\author{
\alignauthor
Mihajlo Grbovic$^\dag$, Vladan Radosavljevic$^\dag$, Nemanja Djuric$^\dag$, \\ Narayan Bhamidipati$^\dag$, Ananth Nagarajan$^\ddag$\\
$^\dag$Yahoo Labs, $^\dag$Yahoo\\ 
\{mihajlo, vladan, nemanja, narayanb, ananth\}@yahoo-inc.com\\
701 First Ave, Sunnyvale, CA, USA\\
}

\maketitle

\begin{abstract}
As one of the leading platforms for creative content, Tumblr offers advertisers a unique way of creating brand identity. Advertisers can tell their story through images, animation, text, music, video, and more, and promote that content by sponsoring it to appear as an advertisement in the streams of Tumblr users. In this paper we present a framework that enabled one of the key targeted advertising components for Tumblr, specifically gender and interest targeting. We describe the main challenges involved in development of the framework, which include creating the ground truth for training gender prediction models, as well as mapping Tumblr content to an interest taxonomy. For purposes of inferring user interests we propose a novel semi-supervised neural language model for categorization of Tumblr content (i.e., post tags and post keywords). The model was trained on a large-scale data set consisting of $6.8$ billion user posts, with very limited amount of categorized keywords, and was shown to have superior performance over the bag-of-words model. We successfully deployed gender and interest targeting capability in Yahoo production systems, delivering inference for users that cover more than $90$\% of daily activities at Tumblr. Online performance results indicate advantages of the proposed approach, where we observed $20\%$ lift in user engagement with sponsored posts as compared to untargeted campaigns.
\end{abstract}


\section{Introduction}

In recent years online social networks have evolved to become an important part of life for online users of all demographic and socio-economic backgrounds. They allow users to easily stay in touch with their friends and family, discuss everyday events, or share their interests with other users with a click of a button. Tumblr is one such social network, representing one of the most popular and fastest growing networks on the web. Hundreds of millions of people around the world come every month to Tumblr to find, follow, and share what they love. The Tumblr network is a gold mine of content, comprising $200$ million blogs on different topics such as travel, sports, and music, where $85$ million user posts are published on a daily basis. This wealth of user-generated data opens a great opportunity for advertisers, allowing them to promote their products through high-quality targeting campaigns to both blog visitors and blog owners \cite{businesshorizons}. 

The standard, prevalent form of advertising on Tumblr is through \textit{sponsored posts} that appear alongside regular posts in the user's {\it dashboard}, a central page for a Tumblr user, displaying newest posts of followed blogs in form of a stream. This form of advertising, in which advertisements camouflage and co-exists with native content in the stream, is often referred to as \textit{native advertising}. Native advertisements are usually aesthetically beautiful and highly engaging, which typically makes them more enjoyable than regular display ads \cite{bonchi12efficient}. Tumblr launched its native advertising product in May of 2012. Since then, the number of advertisers (or brands) on the platform has grown steadily and reached a milestone of $100$ advertisers in April, 2013. Advertising companies on Tumblr include a number of major brands, such as Coca Cola, Converse, and Nestle. Moreover, $8$ of $10$ most valuable brands are advertising on Tumblr\footnote{\url{marketr.tumblr.com}}, while sponsored posts have generated more than $3$ billion paid ad impressions since the launch of the Tumblr advertising product\footnote{\url{www.comscore.com}}. 

However, a huge marketing potential of Tumblr \cite{businesshorizons} has not been fully exploited, due to the fact that targeting against specific interest and demographics audiences, a targeting component that Tumblr was missing, has become industry standard and many advertisers are in need of such solution.

\begin{figure*}[t!]
\centering
\subfloat{\label{fig:undirfig}\includegraphics[width=0.3297\textwidth]{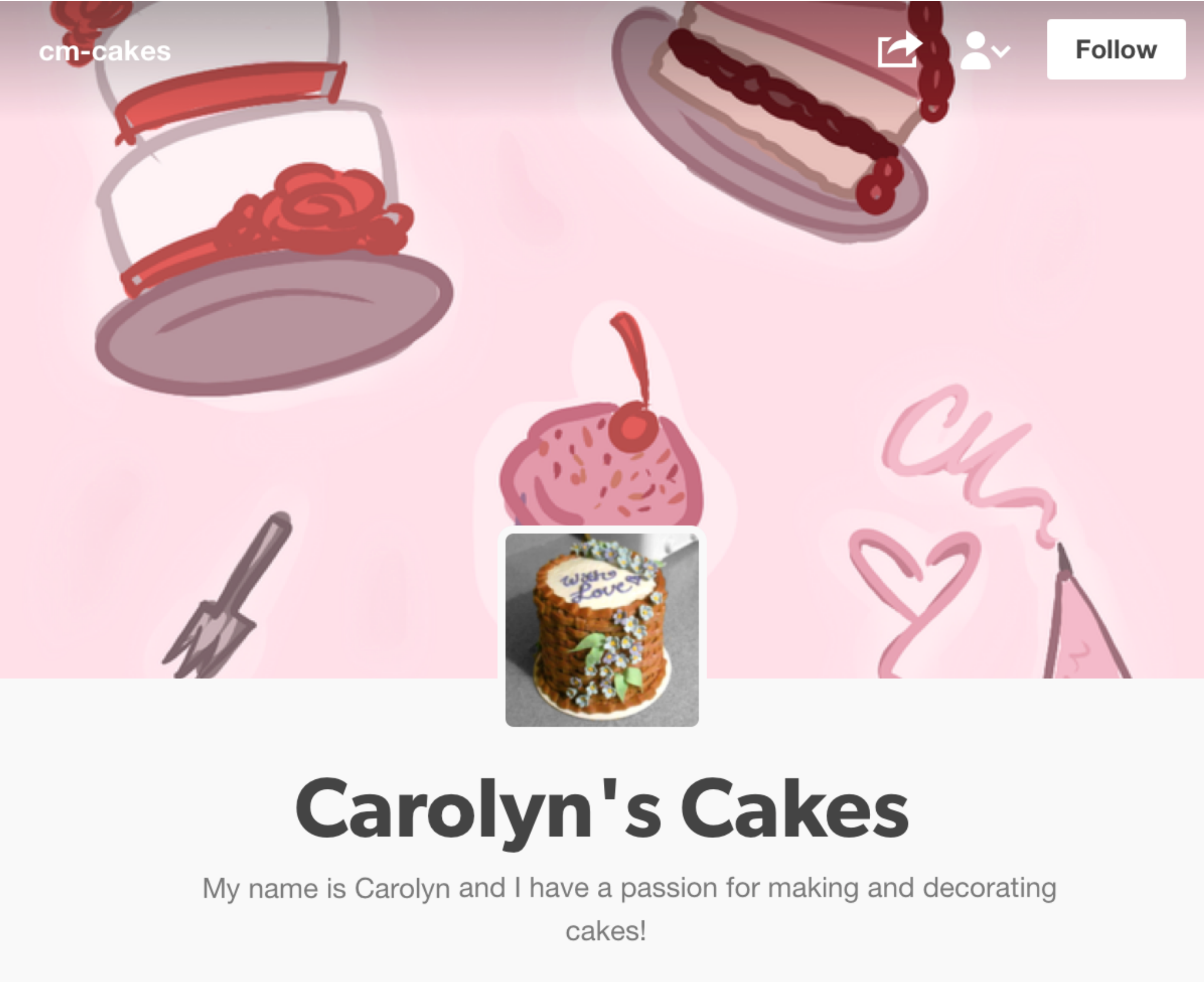}}
\subfloat{\label{fig:dirfig}\includegraphics[width=0.345\textwidth]{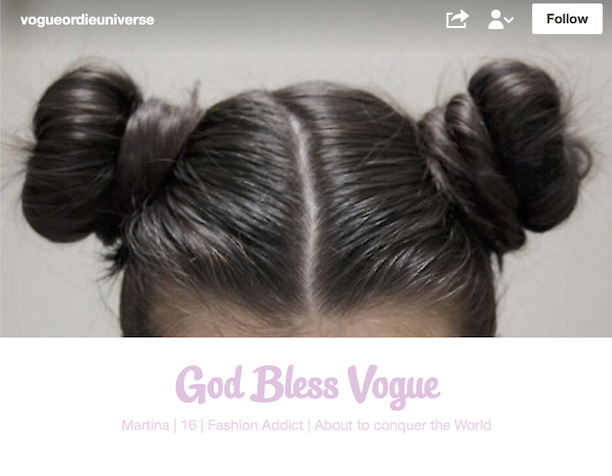}}
\subfloat{\label{fig:dirfig}\includegraphics[width=0.311\textwidth]{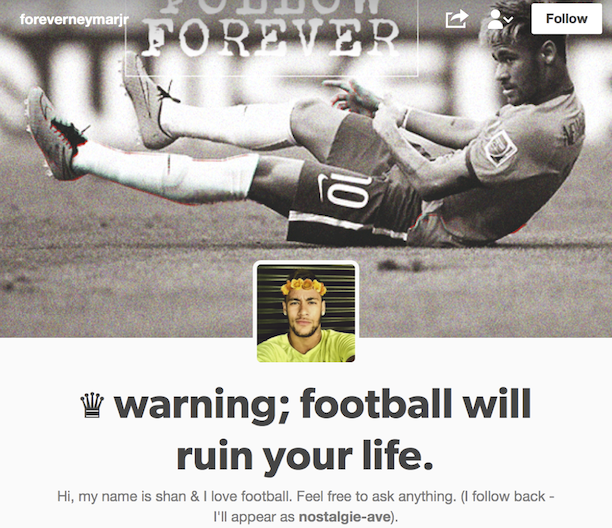}} 
\caption{Examples of blog title (first line, bigger font) and blog description (bottom line, smaller font)}
\label{fig:blog_titles}
\end{figure*}

Building interest targeting products on social and microblogging platforms is an important research topic, discussed previously by several researchers \cite{majumder2013know}. However, due to its distinct characteristics, Tumblr poses novel challenges encountered during the development phase, which we explain in detail in this paper. In particular, content and language used at Tumblr have distinct characteristics that needed to be accounted for during the modeling (e.g., tag ``hp" and word ``hp" have different meanings when they appear in the post, ``Harry Potter" and ``HP company", respectively). Moreover, unlike popular social platform Facebook, which contains large amount of social interactions but limited amount of content, or microblogging platform Twitter, which contain intermediate amount of social interaction and content, Tumblr represents a unique combination of rich and diverse content platform and dynamic social network. To make use of this vast advertising potential, we propose to classify user-generated Tumblr content into a standard multi-level \textit{general-interest taxonomy} \footnote{\url{http://www.iab.net/QAGInitiative/overview/taxonomy}} that advertisers commonly use for defining their targeting campaigns, opening doors to high-quality audience segmentation and modeling for purposes of ad targeting. 
However, inferring categories of users' post is a challenging task, given the huge quantities of unlabeled data being posted every day and very limited amount of labeled data, typically obtained by human editorial efforts. To this end, we propose a novel semi-supervised neural language algorithm, capable of jointly learning embeddings of post keywords, post tags and category representations in the same feature space. The neural model was trained on a large-scale data set comprising of $6.8$ billion posts, with only a fraction of categorized content. 

Targeting pipelines described in this paper are being used to show ads to millions of users daily, and have substantially improved Tumblr's business metrics following the launch. On our path to developing targeting capabilities for Tumblr, we first created user profiles, based on users' Tumblr activities that include publishing blog posts, following other blogs, liking posts, and others. Lastly, we aimed at building and delivering both demographic and interest predictive models based on the created profiles.

We note that the privacy of our users is of critical importance. Therefore, we were constrained on what data we can use. Specifically, user profiles were created solely from data which users share publicly with others, including contents of blog posts, blog title and description, and follow, like and reblog actions. This data is publicly available through Tumblr Firehose data source\footnote{\url{gnip.com/sources/tumblr}}. Other user activities, such as users searches on Tumblr, which blogs they visited and where they clicked, are considered to be sensitive data and were not used in any way for the development of ad targeting models.

\section{Related Work}
Personalization is defined as "the ability to proactively tailor products and product purchasing experiences to tastes of individual consumers based upon their personal and preference information" \cite{chellappa2005personalization}, and it has become an important topic in the recent years. Personalizing online content for individual users may lead to improved user experience and directly translate into financial gains for online businesses \cite{riecken2000personalized}. In addition, personalization fosters stronger bond between customers and companies, and can help in increasing user loyalty and retention \cite{alba1997interactive}. For these reasons it has been recognized as a strategic goal and is a focus of significant research efforts of major internet companies \cite{das2007google,manber2000yahoo}. 

We consider personalization through the domain of ad targeting \cite{essex2009matchmaker}, where the task is to find best matching ads to be displayed for each individual user. This improves user's online experience (as only relevant and interesting ads are shown) and can lead to increased revenue for the advertisers (as users are more likely to click on the ad and make a purchase). Due to its large impact and many open research questions, targeted advertising has garnered significant interest from machine learning community, as witnessed by a large number of recent workshops\footnote{\url{www.targetad-workshop.net}}.  and publications \cite{broder2008computational,majumder2013know}.

One of the basic approaches in ad targeting is to target users with ads based on their demographics, such as age or gender. Historically, this approach has proven to work better than targeting random users. However, while for some products this type of targeting may be sufficient (e.g., women's makeup, women's clothing, man's razors, man's clothing), for others it is not effective enough and more involved profiling of users is required. A popular method in today's ad targeting that addresses this issue is known as interest targeting, where ads are assigned categories, such as ``sports'' or ``travel'', and machine learning models are trained to predict user interest in each of these categories using historical user behavior \cite{ahmed2011scalable,pandey2011learning,tyler2011retrieval}. Typically, a taxonomy is used to decide on the targeting categories, and for each ad category a separate predictive model is trained, able to estimate the probability of an ad click. Then, the models are evaluated on the entire user population, with $N$ users with the highest score selected for ad exposure. In this work we take this approach to develop a ad targeting platform on Tumblr. 

To the best of our knowledge, the Tumblr social network is considered by only a few scientific studies. In \cite{barbieri2014follow,shinrecommending} the authors discuss the problem of blog recommendation, while in \cite{chang2014tumblr} explore Tumblr social norms. However, our work is the first paper that addresses ad targeting at Tumblr.

\section{What is Tumblr?}

Tumblr\footnote{\url{www.tumblr.com}} is one of the most popular blogging services available online, where users can create and share posts with the followers of their blogs. According to data from January 2015\footnote{\url{www.tumblr.com/about}}, there are in total $221.6$ million blogs at Tumblr, that jointly produced over $102.7$ billion blog posts. With large number of users signing up every day, Tumblr is currently the fastest growing social platform\footnote{\url{http://t.co/3txHFRJreJ}}

\subsection{User activities on Tumblr}
To register for a Tumblr account, a valid e-mail address is required, along with a primary username (which will become a part of the blog URL) and a confirmation of age. A Tumblr blog resembles a webpage, with a profile picture, blog title, and blog description appearing on the top (see Figure~\ref{fig:blog_titles}), followed by a stream of blog posts bellow. A first blog created by a registered user is considered his/her primary blog. In addition, a very small portion of users maintains one or more secondary blogs. A Tumblr user is uniquely described by the blog ID of a primary blog, and throughout the paper we will use ``blog" and ``user'' interchangeably. 

\begin{figure}[t]
\centering
{\includegraphics[width=0.32\textwidth]{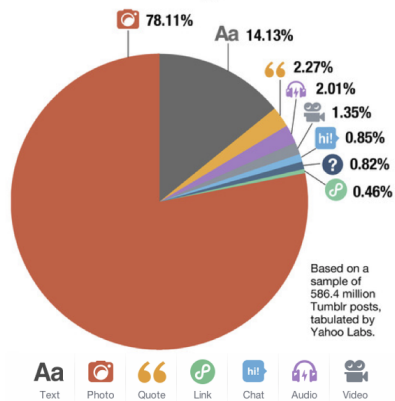}} 
\caption{Distribution of Tumblr post types}
\label{fig:blog_types}
\end{figure}

\begin{figure}[t]
\centering
{\includegraphics[width=0.41\textwidth]{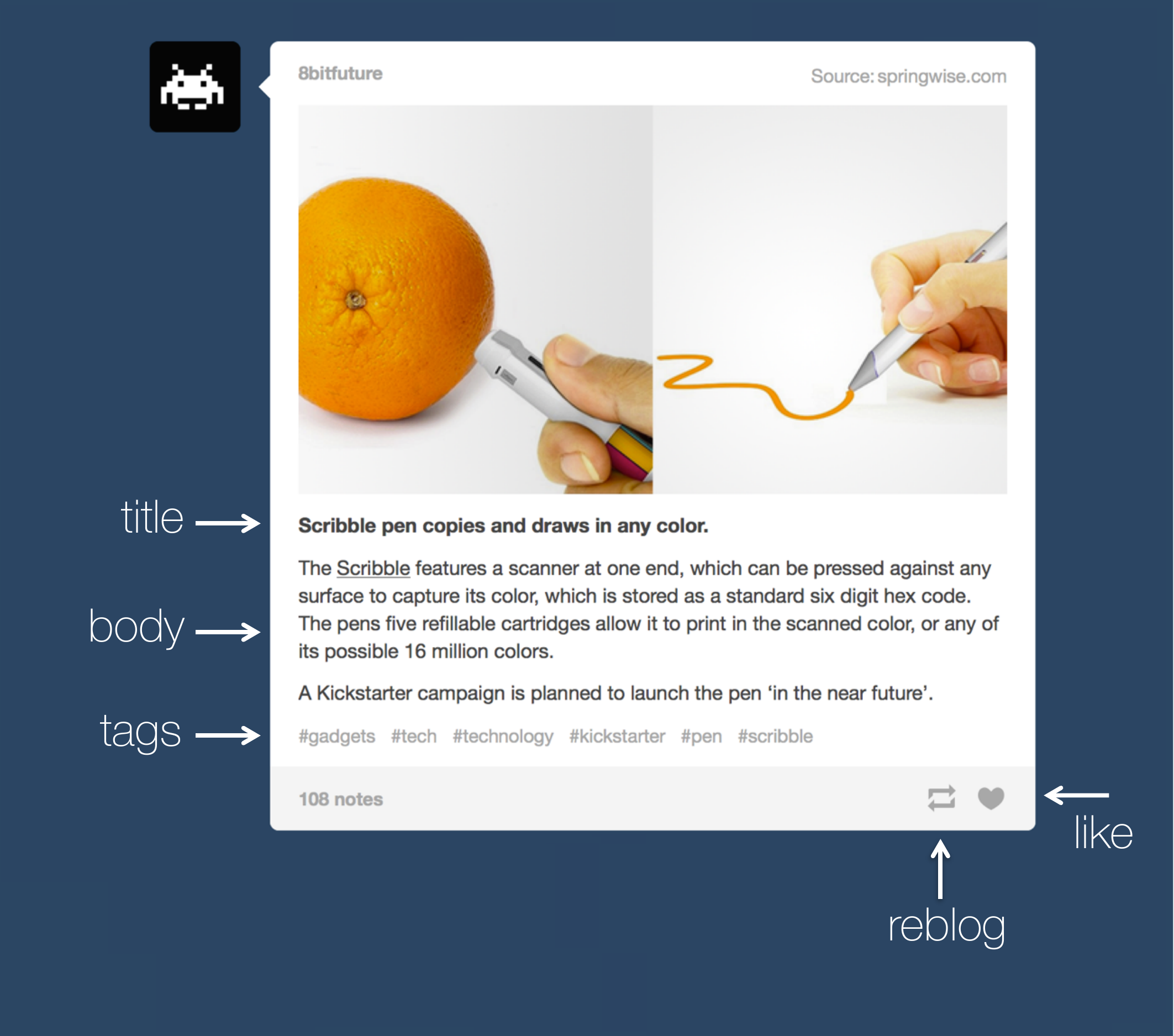}} 
\caption{Example of Tumblr blog post}
\label{fig:blog_post}
\end{figure}

Common user activities of users on Tumblr include the following: 1) creating a post on one's blog; 2) sharing a post created by another blog, called {\it reblogging} (a reblogged post will appear on the user's blog); 3) liking a post by another blog; and 4) following another blog. Similar to Twitter, the follow connections at Tumblr are uni-directional. However, unlike Twitter, users can create longer, richer, and higher quality content in the form of several post types such as text, photo, quote, link, chat, audio, and video. The most popular types of blog posts are photo posts and text posts, and, based on the analysis published in \cite{whatistumblr}, together they cover more than $92\%$ of all posts on Tumblr (see Figure~\ref{fig:blog_types}). Any post type can be annotated with words starting with $\#$ that concisely describe the post and allow for easier browsing and searching (called {\it tags}). Additional metadata that describes a post include photo captions in photo posts, post titles in text posts, and artists names in audio posts. An example post is shown in Figure~\ref{fig:blog_post}. Tags, such as \textit{\#gadgets} or {\it \#tech}, are displayed bellow the photo caption, while the buttons for reblog and like actions are located in the bottom right corner.
Lastly, each user has a {\it dashboard} (i.e., a feed of blog posts published by followed users, ordered in time), with more recent posts appearing at the top.

\begin{figure}[t]
\centering
{\includegraphics[width=0.41\textwidth]{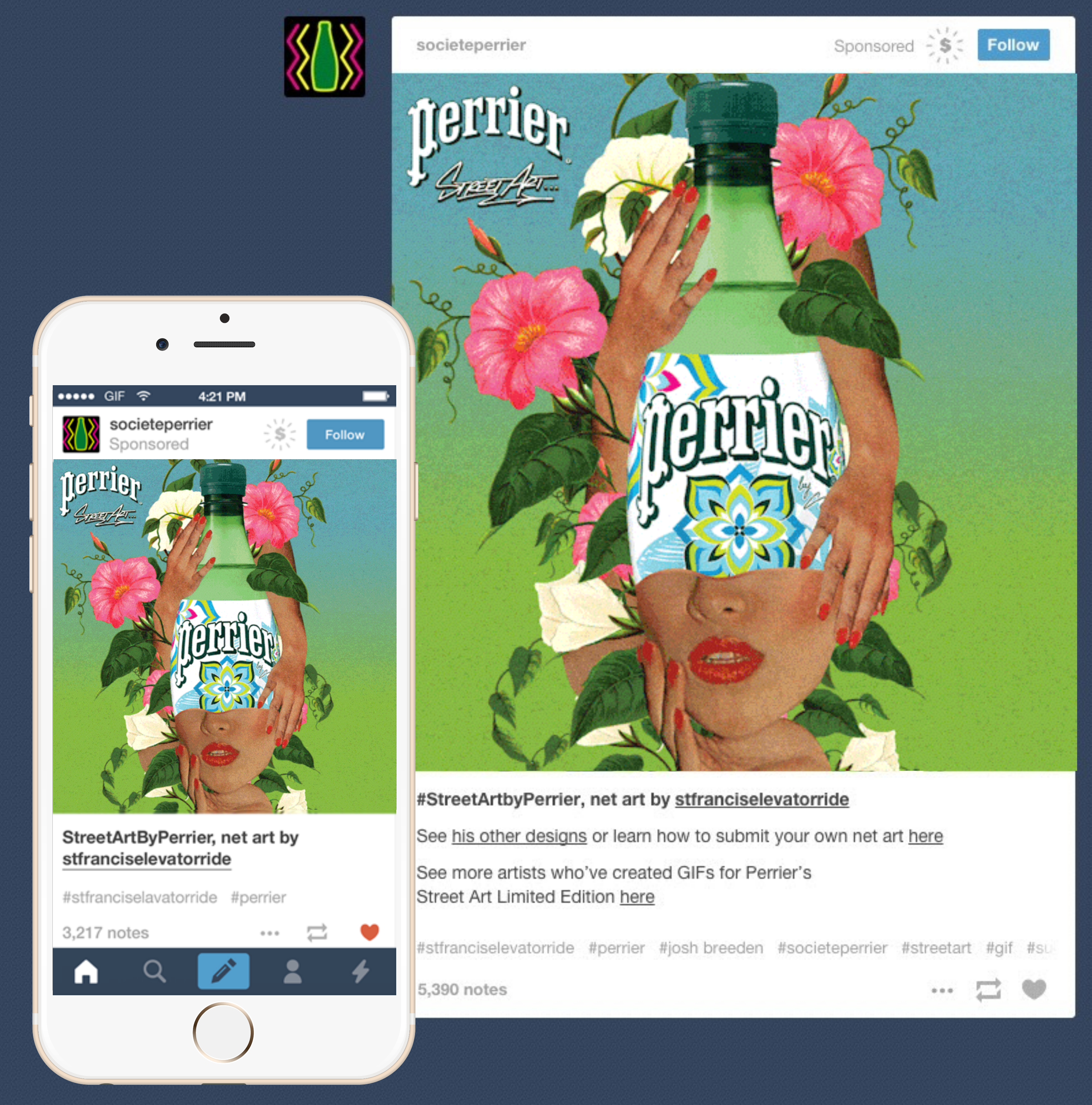}} 
\caption{Example of Tumblr Sponsored post}
\label{fig:sponsored_blog_post}
\end{figure}

\subsection{Advertising at Tumblr}

Advertising at Tumblr is implemented through the mechanism of sponsored (or promoted) posts shown in a user's dashboard. This is similar to how advertising works on Twitter and Facebook. A sponsored post can be a video, an image, or simply a textual post containing an advertising message. In Figure~\ref{fig:sponsored_blog_post} we show an example of a sponsored post and how it appears on web and mobile dashboards. Similarly to organic (or non-promoted) posts, sponsored posts can propagate from user to user in the network by means of reblogs, and can also ``like" the promoted post. Both likes and reblogs can be seen as an explicit form of acceptance or endorsement of the advertising message. Moreover, just like any other posts, sponsored posts are supplemented with notes on who liked and reblogged the post. 

Interestingly, while user-generated, organic posts are reblogged on average $14$ times, sponsored posts are reblogged on average $10{,}000$ times\footnote{http://yhoo.it/1vFfIAc}. We have observed that $40\%$ of engagements with sponsored posts are reblogs, likes, or follows. What is more, every fourth reblog of a sponsored post results in $6$ downstream reblogs from followers, leading to content longevity, and one third of reblogs of sponsored posts occur $30$ days or more after the initial post. 

\section{Tumblr data}
\label{sect:tumblr_data}

In this section we describe the data sources (user activities and post contents) utilized to create user profiles for ad targeting. In particular, user activities included actions such as posts, likes, follows, and reblogs, while post contents included tags, title, and body for text posts, artist names from audio posts, as well as tags and caption for photo posts.

\subsection{Data sources}
Once signed in onto Tumblr, a user can follow other users' blogs. The follow action is one-directional as it does not require the follow back. For the purpose of this study, we collected a sub-graph which contained $96.9$ million unique nodes (i.e., users), $5.1$ billion edges (i.e., follows), out of which $36.4$ million are bi-directional ($18.2$ million pairs of users that follow each other). The data set included more than $26.1$ billion activities on Tumblr. As discussed earlier, activity log is available through a data feed called Firehose.

To create user profiles for targeting, textual contents of all posts were collected, including photo captions, tags, title, and body. In addition, every time a user performs post or reblog activity, Firehose lists the user's blog title and blog description, which we also used to represent a user. As we can see in Figure~\ref{fig:blog_titles}, blog title and description often provide useful information with respect to targeting, such as user's first name, age, and even declared interests (e.g., statements such as ``fashion addict'' or ``I love football'').

\subsection{Keyword extraction}
\label{sec:kw_ext}
In order to improve the representation of user profiles, we propose to extract relevant keywords from available blog information. It is common that certain words appear together more than some others (e.g., words ``credit'' and ``card"), and we aim to capture those bigrams and use them to represent users in addition to separate words. To detect bigrams we use the procedure that counts the unigram and bigram appearances, and for each combination of words $w_i$ and $w_j$ it calculates the following score:
\begin{equation}
\text{score}(w_i , w_j ) = \frac{\text{count}(w_i, w_j )}{\text{count}(w_i) \text{count}(w_j)}.
\end{equation}
Finally, in addition to unigrams, bigrams with score above a certain threshold were chosen to be treated as user keywords, and used to generate rich user profiles.

\subsection{Keyword-based user profiles}
\label{sect:tumblr_profiles}

The available data sources were used to extract user profiles. In particular, we extracted three distinct groups of user-related data: 1) declared; 2) content of posts; and 3) actions. The specific components included in each of data groups are listed in Table~\ref{tab:feat}. From each group we extracted relevant keyword to represent users as described below. 

\begin{table}[t]
\centering
{\footnotesize
\caption{User data extracted from Tumblr Firehose}
\begin{tabular}{l l l }

{\bf Declared} & {\bf Content} & {\bf Actions} \\
\hline
\hline
\rowcolor{lightgray}
blog title & post tags &reblog \\
blog description & photo captions & like \\
\rowcolor{lightgray}
& text post title & follow \\
& text post body & \\
\rowcolor{lightgray}
& audio post artists & \\
\bottomrule
\end{tabular}
\label{tab:feat}
}
\end{table}

{{\bf Declared data}} consist of information which user provided during sign-up, including keywords from blog title and blog description, where keywords were extracted using method in Section~\ref{sec:kw_ext}. To create user profiles we kept the most frequent keywords from blog titles and descriptions, after removing stopwords such as ``a'', ``the'', ``where'', ``in''. We counted the keyword frequency in user's blog title and description, and stored the count and the time stamp of the as a part of the user profile.

{{\bf Content features}} were formed from textual contents of posts which user either created or reblogged. The main content feature types included: 1) post tags; 2) keywords from post title and body; 3) keywords from photo post captions, and 4) artist names from audio posts. 
Tags in posts were not tokenized, instead they were used in the form they appear, e.g., tag ``food for a vegan'' was one keyword. On the other hand, to extract keywords from text appearing in text post content we again used the method from Section~\ref{sec:kw_ext}. We kept only the most frequently occurring keywords, excluding stopwords. In addition, we used the most popular artist names as keywords. In this way, we collected several millions of distinct keywords that were used to obtain rich representation of user profiles.
To illustrate content keyword extraction from our dataset, consider that user $u_i$ at time stamp $t$ used tag {\it \#hp} five times and tag {\it \#nba} eight times, keyword {\it football} two times in post titles and posted an audio post with song from artist {\it Shakira} ten times, then the resulting user profile would be: $u_i = \{tag:\{\#hp, t:5, \#nba:8\}, title:\{football, t:2\}, artist:\{shakira, t:10\}\}$.

{{\bf Action features}} include follows, likes, and reblogs. If a user $u_i$ follows user $u_j$ at time stamp $t$, we created an indicator feature $follows:\{j,t:1\}$ and add it to the $u_i$'s user profile. Similarly, if user $u_i$ likes or reblogs user's $u_j$ post, we created a count feature that keeps record of the number of likes $m$ and reblogs $n$, as $likes:\{j,t:m\}$, $reblogs:\{j,t:n\}$, respectively, and update the user profile accordingly. Furthermore, in order to help enrich profiles for users who do not post content but only follow and like other posts, we identified frequent bloggers in each interest category. To identify frequent bloggers for category $k$ we ranked users by the the total number of categorized activities and retained $5\%$ of users with the highest feature value. Then, for each interest category from the taxonomy we created an additional feature in user representation vector that counts how many frequent bloggers from that category a user is following. 

The described approach resulted in user profiles for a total of $81.8$ million users. The total number of unique keywords was $1.4$ million, and an average user had $379.9$ non-zero features. 
Most of the keywords described above are represented as either binary indicators or counts of occurrences. To handle large counts, we normalize the numerical data at a user level by applying log transformation: assuming that the count is $x$, we replace the count by the value $\log(1+x)$. 

\section{Interest Prediction}
\label{sed:interest}
The goal of our work is to infer user demographics (described in the following section) and identify user groups with interest in certain topics, such as music, travel, cooking, or books, to allow advertisers to target segmented Tumblr audiences. As the topics may be defined at various levels of granularity, to avoid sparsity problems while still provoding useful and actionable interest categories, user interests are often classified into pre-determined hierarchical interest taxonomy that the advertisers commonly use. 
However, to be able to create effective user interest classifiers one requires sufficient amount of labeled data.
Yet, for the problem of the scale of Tumblr interest prediction this can be a daunting task for human editors. For that reason we propose a novel semi-supervised classification approach, based on recently proposed word2vec model \cite{mikolov2013distributed}, that efficiently and seamlessly makes use of large amounts of unlabeled and limited amount of editorially labeled data for learning effective interest classifiers.

\begin{figure}[t!]
\centering
\label{fig:skip_gram_reg_un}
\includegraphics[width=0.276\textwidth]{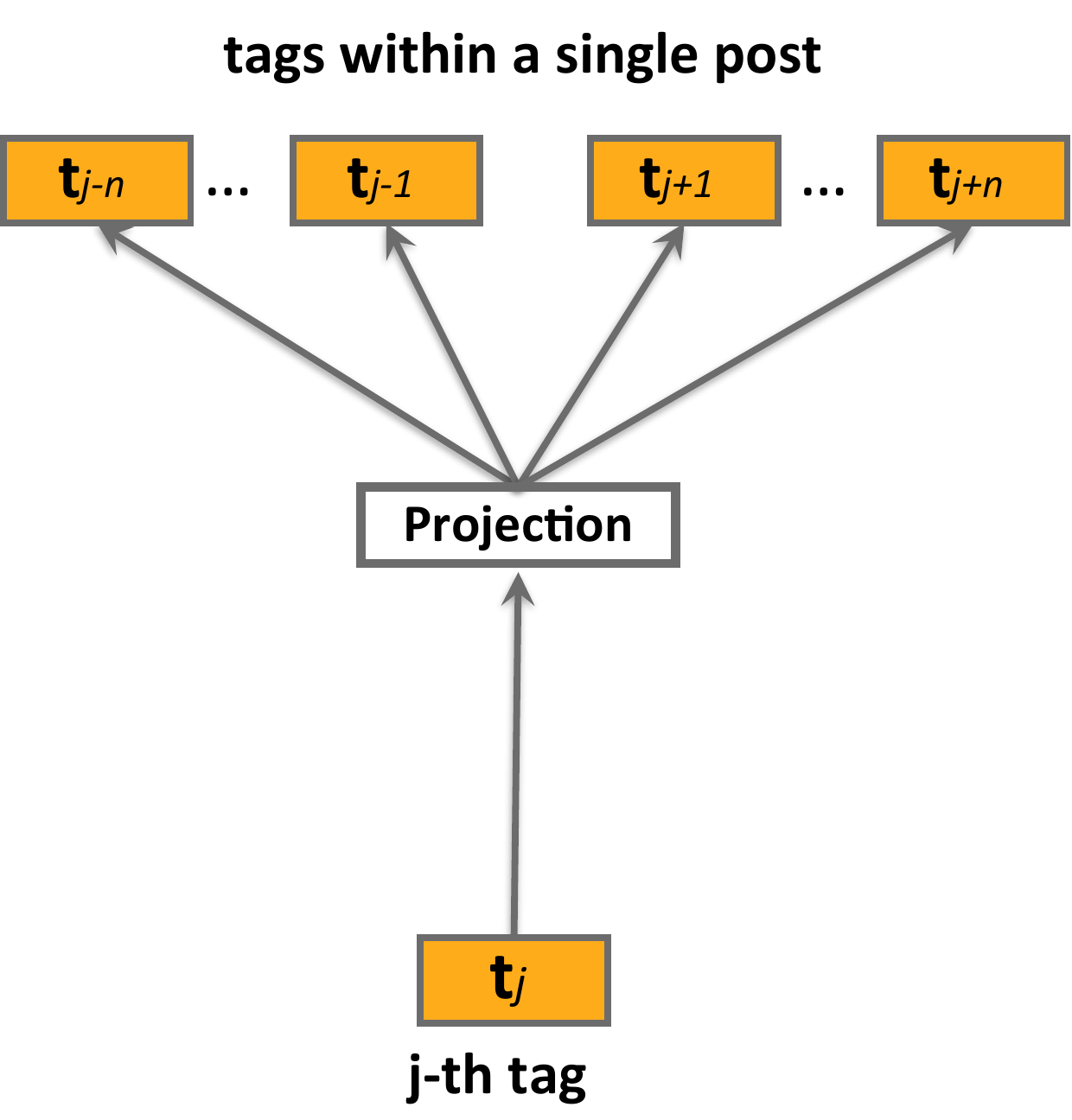}
\caption{Unsupervised skip-gram model}
\label{fig:skipgram_un}
\end{figure}

\begin{figure}[t!]
\centering
\label{fig:skip_gram_reg_ss}
\includegraphics[width=0.35\textwidth]{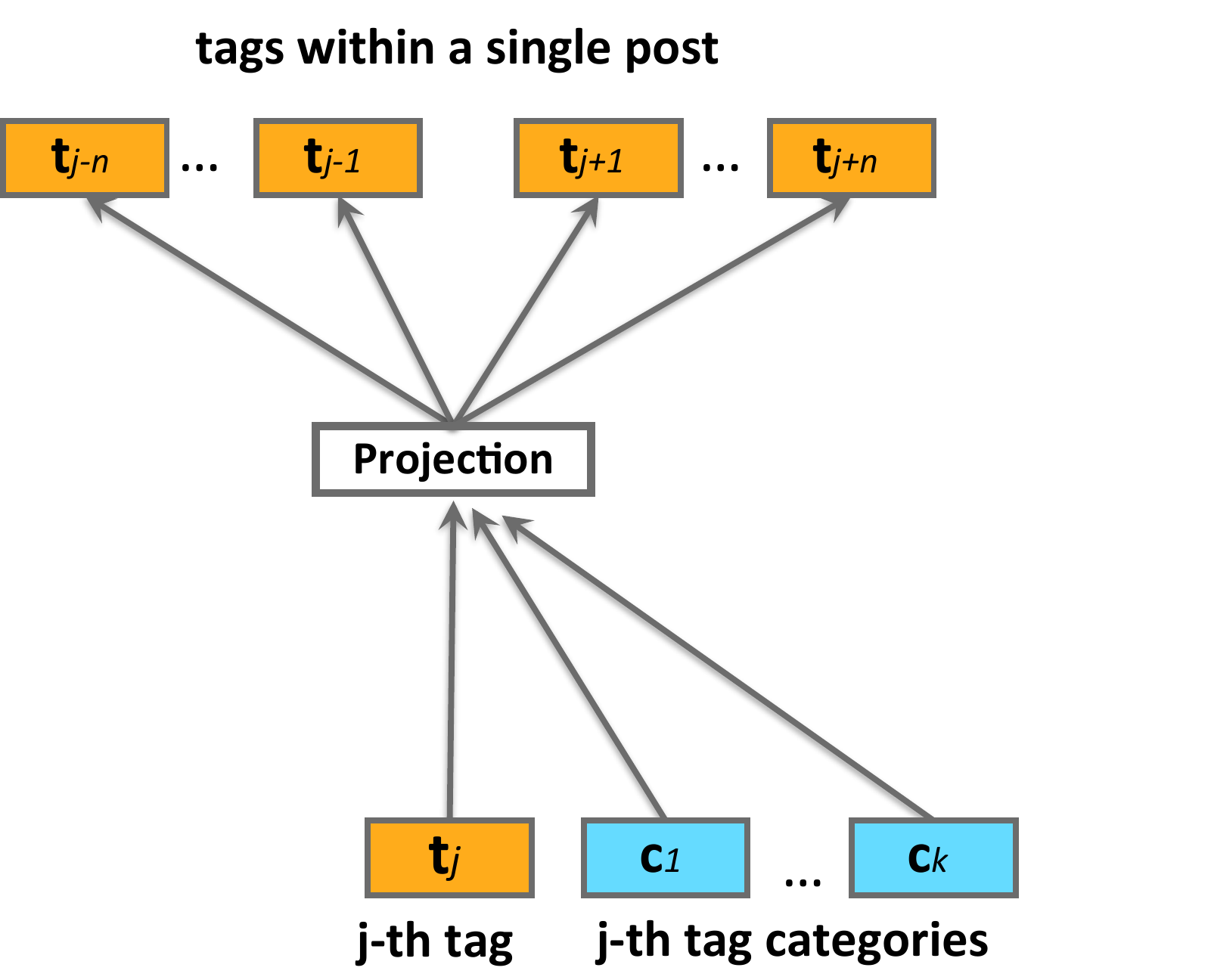}
\caption{Semi-supervised skip-gram model}
\label{fig:skipgram_ss}
\end{figure}

\subsection{User interest taxonomy}
We decided to classify keywords into General Interest Taxonomy (GIT), used by the Yahoo Gemini advertising platform for native advertising\footnote{\url{gemini.yahoo.com}}. The GIT is carefully derived based on Interactive Advertising Bureau (IAB) taxonomy recommendations in order to meet advertiser needs and protect Yahoo's interests. The GIT has a two-level hierarchical structure such that advertisers can adjust the audience reach by utilizing broader or narrower interest categories. The top level of the taxonomy contains $23$ nodes (e.g., ``Automotive", ``Business", ``Pets", ``Travel"), while the second level contains $130$ nodes which represent more precise interests (e.g., ``Automotive/SUV", ``Automotive/Luxury", ``Pets/Dogs").

\subsection{Proposed semi-supervised classification}
\label{sec:semi}
In this section we present a novel classification approach based on recently proposed skip-gram model \cite{mikolov2013distributed}, used to categorize keywords into the GIT taxonomy. For conciseness, we describe the proposed model assuming that it is applied to tag categorization. However, we used the same methodology for categorization of keywords originating from blog title, description, and text, audio, and image posts. 

We consider the task of tag classification, where the goal is to classify tags into a pre-defined taxonomy of interest categories. In order to address this problem, we propose to learn tag representation in a low-dimensional space using neural language models applied on historical Tumblr posts.
Let us assume we are given $N$ posts. In post logs found in Firehose, every post $p_i$ is recorded along with the tags $t_{ij}$, $j=1...M$. We collected data in the form $p_i = \{t_{ij}, j=1...M_i\}$, where $M_i$ represents the number of tags in the $i$-th post. Given data set $\mathcal{D} = \cup_{i=1}^{N} p_i$, the objective is to find representation of tags such that semantically similar tags are nearby in the representation space. For this purpose we extend ideas originating from recently proposed language models, as described in the remainder of the section.

{\bf The skip-gram (SG) model} involves learning representations of tags in a low-dimensional space from post logs in an unsupervised fashion, by using a notion of a blog post as a ``sentence'' and the tags within the post as ``words'', borrowing the terminology from Natural Language Processing (NLP) domain (see Figure \ref{fig:skipgram_un}). Tag representations using the skip-gram model \cite{mikolov2013distributed} are learned by maximizing the objective function over the entire set $\mathcal{D}$ of blog posts, defined as
\begin{equation} \label{cx2v_obj}
\mathcal{L} = \sum_{p \in \mathcal{D}} \sum_{t_j \in p} \sum_{-n\le m\le n, m\ne 0} \log \P(t_{j+m}|t_j).
\end{equation} 
Probability $\P(t_{j+m}|t_j)$ of observing a neighboring tag $t_{j+m}$ given the current tag $t_j$ is defined using the soft-max,
\begin{equation}\label{cx2v_2}
\P(t_{j+m}|t_j) = \frac{\exp(\mathbf{v}_{t_j}^\top \mathbf{v}_{t_{j+m}}^\prime)}{\sum_{t=1}^T \exp(\mathbf{v}_{t_j}^\top \mathbf{v}_{t}^\prime)},
\end{equation}
where $\mathbf{v}_{t}$ and $\mathbf{v}_{t}^\prime$ are the input and output vector representations of tag $t$ of user-specified dimensionality $d$, $n$ defines the length of the context for tag sequences, and $T$ is the number of unique tags in the vocabulary. From equation \eqref{cx2v_2} we see that tags that often co-occur and tags with similar contexts (i.e., with similar neighboring tags) will have similar vector representations as learned by the word2vec model.

{\bf The semi-supervised skip-gram (SS-SG) model} assumes that some tags are labeled with categories from GIT taxonomy. Then, we introduce a dummy category vector for each node of the taxonomy, and leverage tag contexts in blog posts to jointly learn tag vectors and category vectors in the same feature space. Given such setup, after learning the representations every tag from the vocabulary can be categorized by simply looking up the closest category vector in the joint embedding space. 

\begin{table}[t]
\centering
{
\caption{Examples of categorized tags}
\begin{tabular}{ l l }
{\bf Tag} & {\bf Category} \\
\hline
\rowcolor{lightgray}
music & Arts and Entertainment/Music \\
fashion & Style and Fashion \\
\rowcolor{lightgray}
song & Arts and Entertainment/Music \\
art & Arts and Entertainment \\
\rowcolor{lightgray}
disney & Arts and Entertainment/Movies \\
style & Style and Fashion \\
\rowcolor{lightgray}
photography & Hobbies/Photography \\
teen wolf & Arts and Entertainment/TV \\
\rowcolor{lightgray}
food & Food and Drink \\
\bottomrule
\end{tabular}
\label{tab:tags}
}
\end{table}

\begin{figure}[t!]
\centering
\subfloat{\label{fig:tag_neighbors_a}\includegraphics[width=0.235\textwidth]{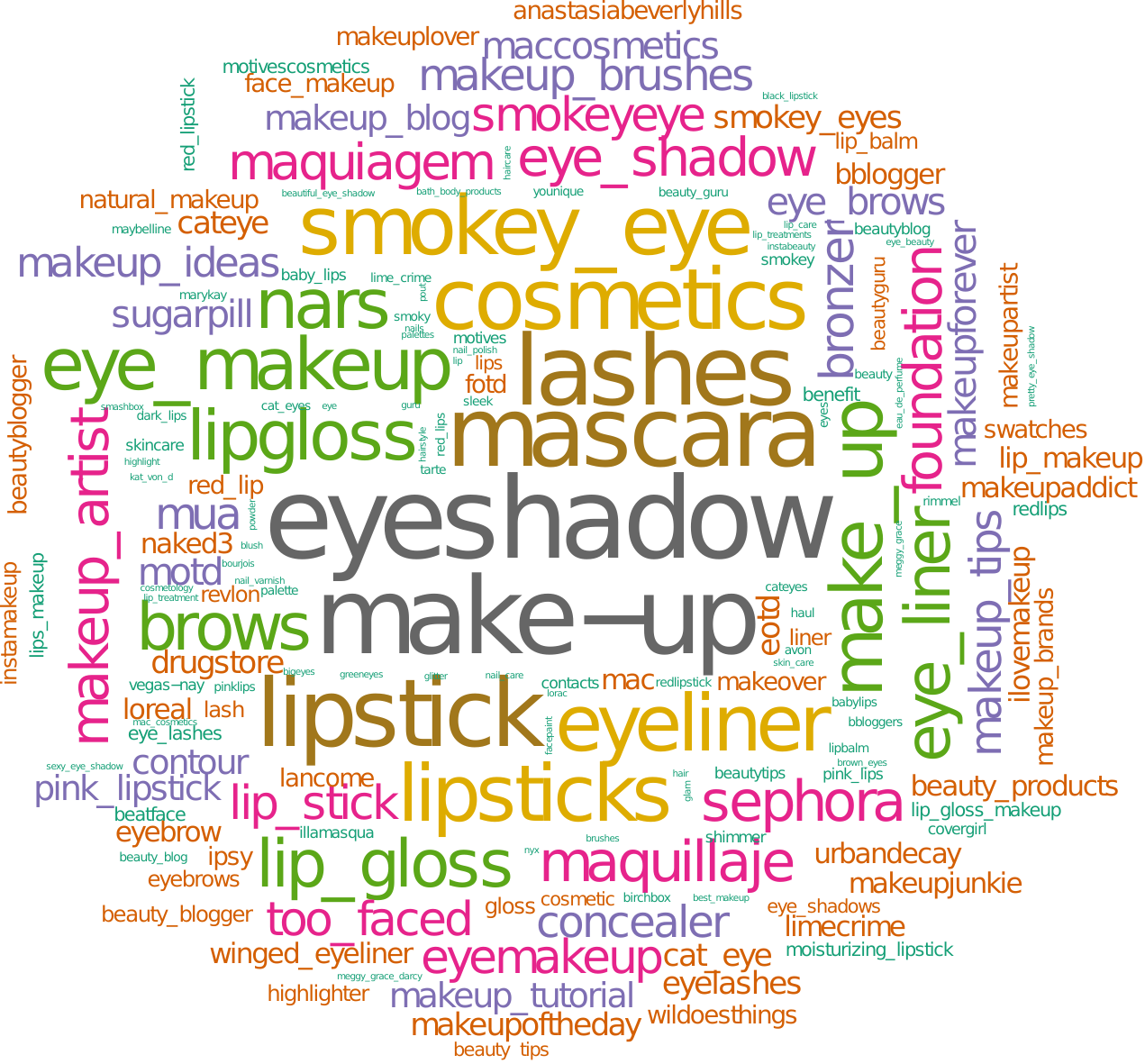}}
\subfloat{\label{fig:tag_neighbors_b}\includegraphics[width=0.24\textwidth]{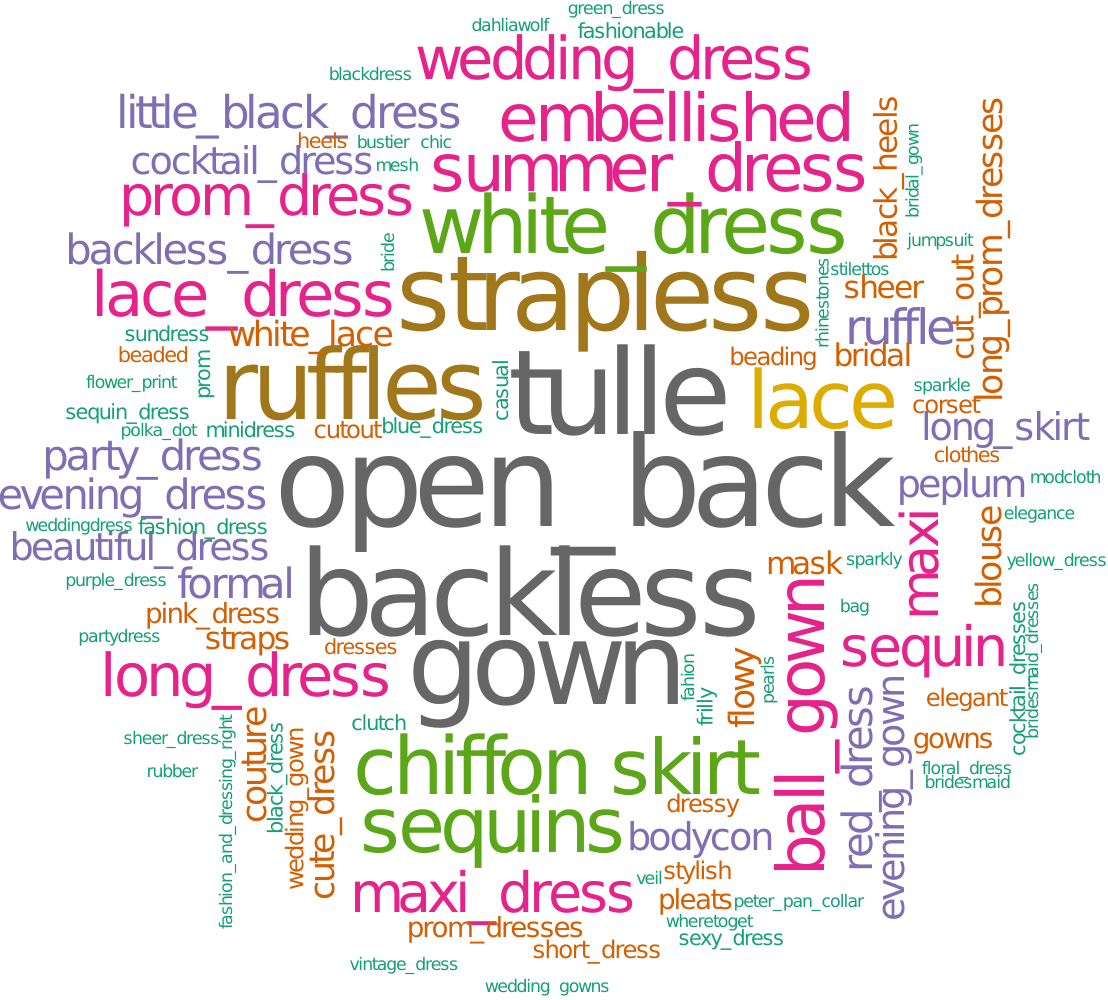}} 
\caption{Nearest neighbors of tags: a) {\it \#makeup}; b) {\it \#dress}} 
\label{fig:tag_neighbors}
\end{figure}

Specifically, given the labeled tags, we extend $\mathcal{D}$ to obtain data set $\mathcal{D}_{ss}$ where categories are imputed into post `sentences'' $p_i$, where available. In particular, labeled tags are accompanied by assigned categories, and every time a vector of labeled central tag $t_j$ is updated to predict the surrounding tags, vectors of categories assigned to $t_j$ are updated as well. More formally, assuming central tag $t_j$ is labeled with $C_j$ of $C$ categories in total, $\zeta_j=\{c_1, \ldots, c_{C_j}\}$, the semi-supervised skip-gram learns tag and category representations by maximizing the following objective function,
\begin{equation} \label{category_obj}
\sum_{p \in \mathcal{D}_{ss}} \sum_{t_j \in p} \sum_{-n\le m\le n, m\ne 0} \Big( \log \P(t_{j+m}|t_j) + \sum_{c \in \zeta_j} \log \P(t_{i+m}|c) \Big).
\end{equation} 
Probability $\P(t_{i+m}|c)$ of observing tag $t_{i+m}$ given label $c$ of the current tag $t_j$  is defined using the soft-max,
\begin{equation}\label{cx2v_prq}
\P(t_{i+m}|c) = \frac{\exp(\mathbf{v}_{c}^\top \mathbf{v}_{t_{i+m}}^\prime)}{\sum_{t=1}^T \exp(\mathbf{v}_{c}^\top \mathbf{v}_{t}^\prime)}.
\end{equation}
This procedure allowes us to seamlessly incorporate labeled and unlabeled data, and learn tag and category vectors in the joint embedding space. Then, classification of tags amounts to simple nearest-neighbor search among the category vectors. In Figure \ref{fig:skipgram_ss} we show the graphical representation of the semi-supervised skip-gram model.

\subsubsection{Training}
The models are optimized using stochastic gradient ascent, suitable for large-scale problems. However, computation of gradients $\nabla \mathcal{L}$ in \eqref{cx2v_obj} is proportional to the vocabulary size $T$, which may be computationally expensive in practical tasks as $T$ could easily reach several million tags. As an alternative, we used negative sampling approach \cite{mikolov2013distributed}, which significantly reduces the computational complexity. 

\begin{figure}[t]
\centering
{\includegraphics[width=0.40\textwidth]{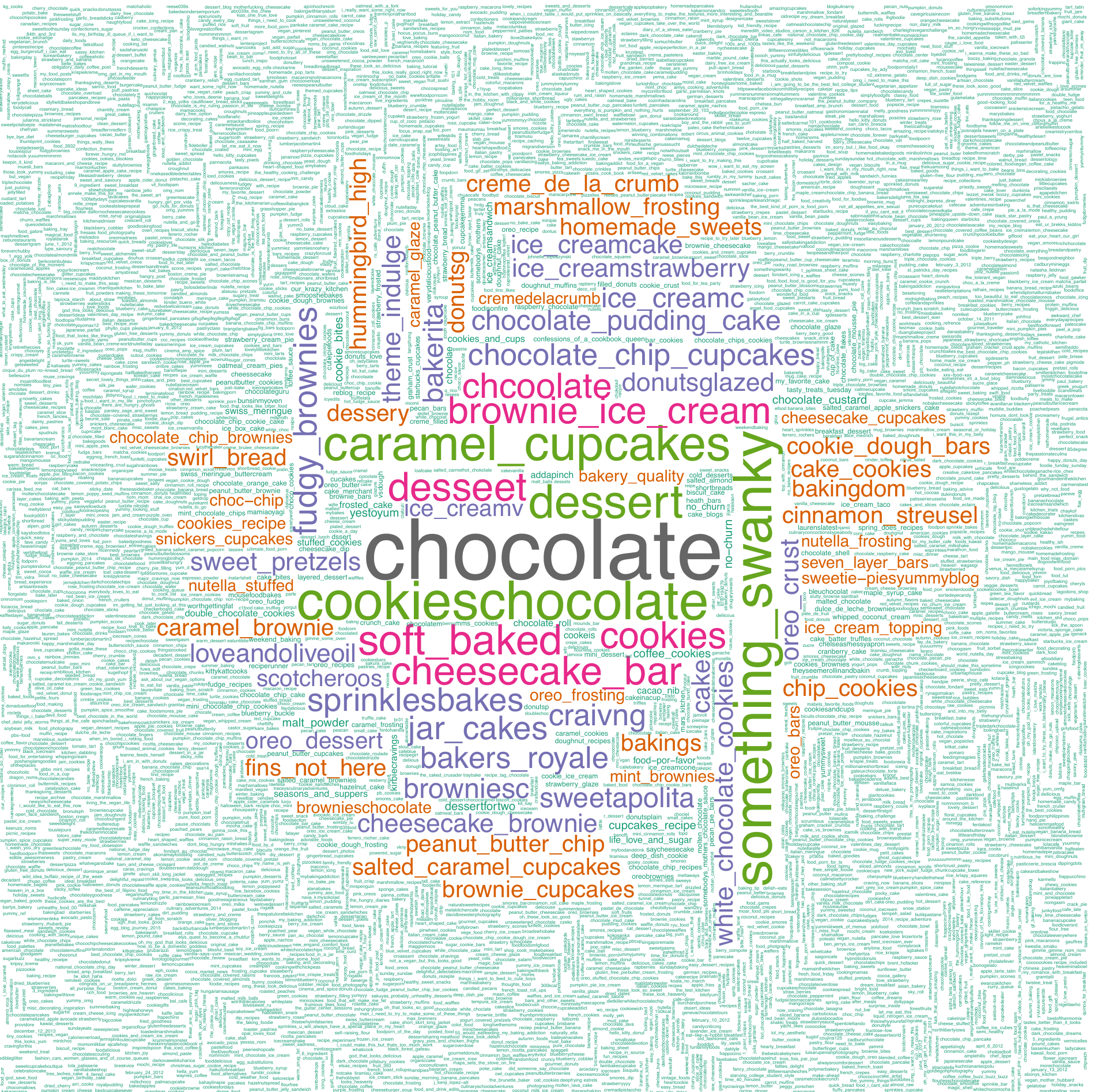}} 
\caption{Nearest tags to ``Food and Drink/Desserts"}
\label{fig:blog_food}
\end{figure}

\begin{figure}[t]
\centering
{\includegraphics[width=0.40\textwidth]{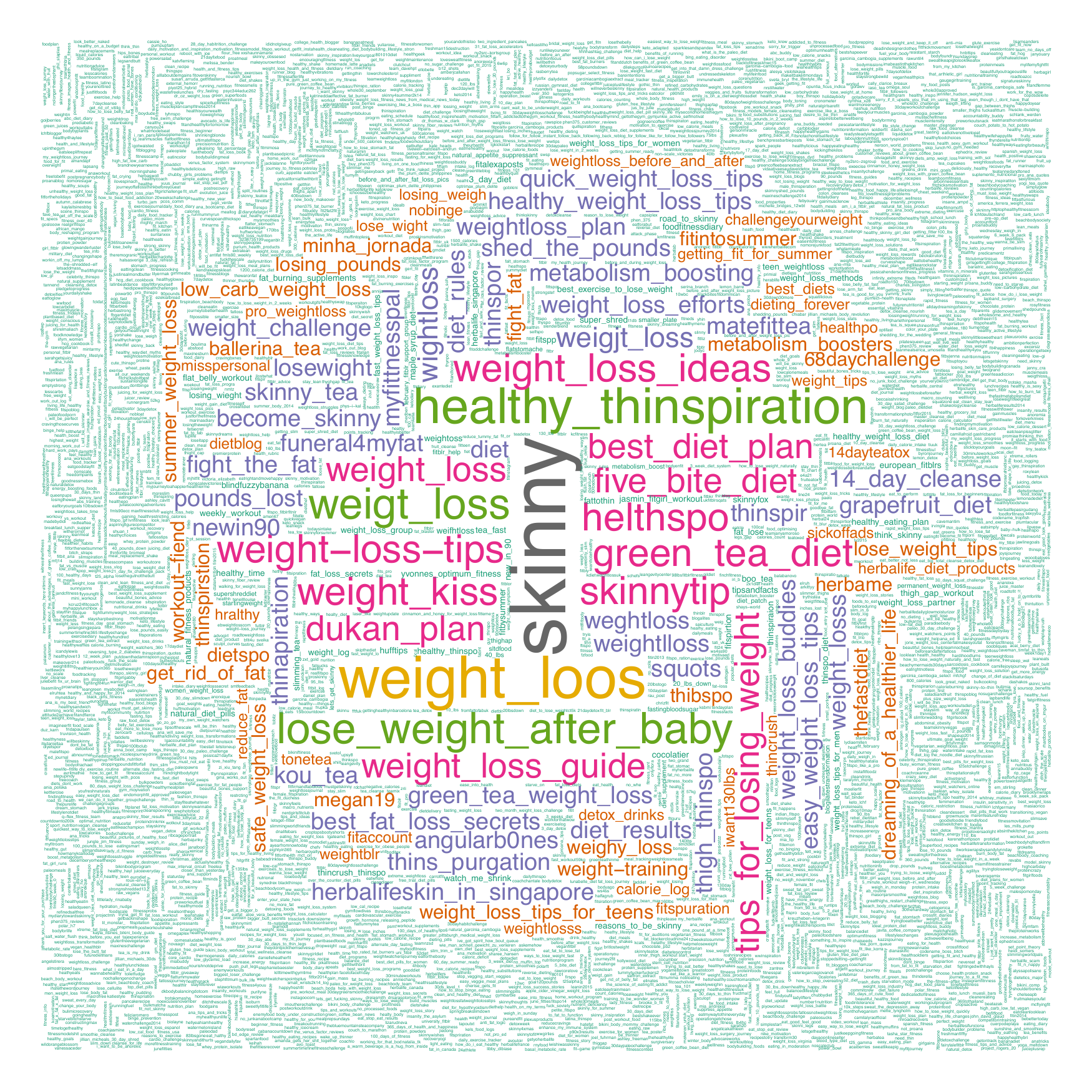}} 
\caption{Nearest tags to ``Health and Fitness/Weight Loss"}
\label{fig:blog_fit}
\end{figure}

Data set used during model training comprised $6.8$ billion posts that contained tags. To collect category labels for some tags, we sorted the tags in the decreasing order of popularity and the editors labeled the top ones with one or more categories. Following this process we obtained $8{,}400$ categorized tags, which covered $10\%$ of the entire data set. We show examples of categorized tags in Table~\ref{tab:tags}.

The models were trained using a machine with $96$GB of RAM memory and 24 cores. Dimensionality of the embedding space was set to $d=300$, context neighborhood size was set to $n=5$. Finally, we used $10$ negative samples in each vector update for negative sampling. Similarly to the approach in \cite{mikolov2013distributed}, most frequent tags were sub-sampled during training.

\subsubsection{Inference}
When the vector representations of all tags are learned, we can find similar tags for a given tag by straightforward $k$-nearest neighbor ($k$-NN) searches in the low-dimensional representation space. We use cosine distance \cite{mikolov2013distributed} as a measure of similarity. To illustrate usefulness of our approach, examples of similar tags to tag {\it \#makeup} and {\it \#dress} are shown in Figure~\ref{fig:tag_neighbors}, where we see that semantically related tags are grouped in the same part of the embedding space.

\begin{table}
\caption{Precision and recall of categorization methods}
\label{tbl:results_bow}
\begin{center}
\begin{tabular}{lcc}
\rule{0pt}{2.5ex}{\bf Method} & {\bf Precision} & {\bf Recall} \\
\hline \hline
\rowcolor{lightgray}
LR-SG & 0.71 & 0.65 \\ 
$k$-NN-SG & 0.82 & 0.62 \\
\rowcolor{lightgray}
SS-SG & {\bf 0.85} & {\bf 0.63} \\
\bottomrule
\end{tabular}
\end{center}
\end{table}

Similarly, we can find the most likely category for any tag by searching for the nearest neighbors in the subset of category vectors. To produce highly confident tags in each category, we calculated cosine distance to each of the tags in the vocabulary and retrieved the tags with cosine distance above $0.7$. This threshold was found by editorial evaluation of the results. In total, more than $380{,}000$ tags were confidently categorized into one or more categories. We show examples of categorized tags for categories ``Food and Drink/Desserts and Baking" and ``Health and Fitness/Weight Loss" in Figures \ref{fig:blog_food} and \ref{fig:blog_fit}, respectively.

{\bf Demonstration video} of our tag categorization tool is available online at \url{http://youtu.be/ygn5oUBydfM}.

\begin{table*}[t]
\centering
{\scriptsize
\caption{Examples of interest inference based on enriched user profiles}
\begin{tabular}{ l l l }
{\bf User} & {\bf Inferred interest} & {\bf Original keyword-based profile} \\
\hline
\rowcolor{lightgray}
user 1 & Arts and Entertainment/Movies & tag:\{spoilers:30, shrek:18, hercules:12, cinderella:3, hobbit:123, hulk:21, pokemon:7, thor:58, \ldots\\
\rowcolor{lightgray}
& & disney:500, tarzan:8, marvel:385, wolverine:21, twilight:2, pixar:87, godzilla:1, x-men:53, \ldots \\
\rowcolor{lightgray}
& & pocahontas:4, avengers:134\}  \\ 
\rowcolor{lightgray}
& & txt:\{aladdin:28, batman:10, bambi:12, movies:100\} \\ 
\rowcolor{lightgray}
& & desc:\{oscar:1, animation:12, comedy:1, movie:1, dvd:1\}  \\
\hline
user 2 & Style and Fashion & tag:\{ womensfashion:110, curls:6, fashiondiaries:133, redhair:2, menswear:125, chanel:4 \ldots \\
& &  springfashion:50, style:132, streetstyle:132, hairstylist:134, dapper:3, mensfashion:124\}\\
& & txt:\{fashion:108\}\\
\hline
\rowcolor{lightgray}
user 3 & Food and Drink & tag:\{food:11, dessert:4 , soup:1, brunch:1, fruit:2, chicken:3, smoothie:1, cake:2, breakfast:2, \ldots \\
\rowcolor{lightgray}
& &  ginger:2, salad:5, avocado:1\} \\
\rowcolor{lightgray}
& & txt:\{food:16, meals:6\} \\ 
\hline
user 4 & Home and Garden & tag:\{daisies:2, kitchen:20, chair:3, art:81, outdoor:20, chandelier:12, lamp:8, window:2, bath:1 \ldots \\ 
& & floral:17, home:3, wildflowers:1, flowers:102,  interior:201, tree:1, flower:49, table:1, stairs:2, \ldots\\ 
& & bedroom:56, wood:2, bathroom:26\} \\
& & txt:\{garden:32, interior:17, home:41\} \\
\hline
\rowcolor{lightgray}
user 5 & Automotive/Motorcycles & tag:\{cars:24, ride:9, vehicle:22, riding:8, road:18\} \\
\rowcolor{lightgray} 
& & txt:\{bike:8, motorcycle:10, riding:5, ride:9, road:10, vehicle:18, bikes:6, bicycle:2, scooter:1\} \\
\bottomrule
\end{tabular}
\label{tab:infres}
}
\end{table*}

\subsubsection{Evaluation}
To quantify the benefits of our approach, we evaluated the method by excluding random $1{,}000$ tags from the editorially labeled set, and training the model using the remaining $7{,}400$ labeled tags. We compared the SS-SG classification to the state-of-the-art logistic regression (LR) and $k$-NN, trained on the vectors learned by the original SG model. For LR classification (we refer to the method as LR-SG) we trained one classifier per interest category, while in $k$-NN (we refer to the method as $k$-NN-SG) for each test tag we found $K=50$ nearest categorized neighbors and predicted the category that appeared more than $10$ times.
We report results following 5-fold cross-validation in Table \ref{tbl:results_bow}. The results indicate that classification based on our approach achieves higher precision than the competing methods, while at the same time maintaining competitive recall measure.

\label{sed:interest}
\begin{table}[t]
\centering
{
\caption{Language differences in post tags and post text}
\begin{tabular}{ l l }
{\bf tag "hp" neighbors} & {\bf word "hp" neighbors } \\
\hline
\rowcolor{lightgray}
harry potter & hewlett packard \\
hp movies & hp.com \\
\rowcolor{lightgray}
hp books & hp computers \\
hp book quotes & hp company \\
\rowcolor{lightgray}
harry potter facts & dell computers \\
hogwarts & hp printers \\
\bottomrule
\end{tabular}
\label{tab:lang}
}
\end{table}

\subsubsection{Model extensions}

To be able to map more of Tumblr content to GIT taxonomy, we trained two more semi-supervised skip-gram models, for: 1) keywords from post title and body and 2) keywords from blog title and description.
To train the models, we followed a similar procedure as before. Editors provided $4{,}700$ categorized keywords, used to form training data sets for SS-SG model learning. Post keyword vectors were trained using a data set comprising $N=6.8$ billion posts, while blog title and description keyword vectors were trained using $N=37.1$ million blogs.
These models were trained separately because of language differences in these three domains, between used language in post tags, post text, and blog title and description text. To justify our claim, in Table \ref{tab:lang} we show nearest neighbors of tag ``hp'' from tag SS-SG model and word ``hp'' from post text SS-SG model. As we can see in the table, ``hp'' has two different meanings in post tags and post text domains, referring to ``Harry Potter" and ``Hewlett-Packard", respectively.

To find the most confident keywords for each category, we calculated cosine distances between category vectors and all keyword vectors from the vocabulary. We retrieved $184{,}000$ text blog keywords with cosine distance of $0.7$ or higher. We repeated the same procedure for keywords from blog title and description, resulting in $173{,}000$ categorized keywords.

\subsection{Forming interest segments}

The goal of the task of interest prediction is to identify groups of users with interest in certain topics, such as music, travel, cooking, or books, to allow advertisers to target Tumblr audience by interests. In the following we describe a method for predicting user interests used in this study.

In particular, after we obtained user profiles with categorized high-confidence tags and keywords, the interest score for user $u_i$ in the $k$-th category at time $t$ was calculated as
\begin{equation}\label{intensity_cat}
u_{i, cat=k}^t = \sum_{act \in \mathcal{A}_i} \sum_{kw \in act} \alpha^{(t-t_{act})} ~w_{kw} ~I(kw \text{ is of class }k),
\end{equation}
where $\mathcal{A}_i$ is a set of all activities by user $u_i$, $w_{kw}$ is the value of the keyword feature (e.g., if the post contains two  mentioned of keyword ``shakira" then the value is equal to 2, as explained in Section \ref{sec:kw_ext}), and indicator function $I(\cdot)$ returns 1 is the keyword extracted from a activity is of class $k$, and 0 otherwise. In addition, we used the time stamp of the activity $t_{act}$ to exponentially decay less recent activities to account for passing interest (we used $\alpha=0.99$ in our experiments). Note that the set $\mathcal{A}_i$, in addition to user's original content, also included posts reblogged by user $u_i$.

The value of $u_{i, cat=k}^t$ represents an exponentially time-decayed count of all activities in the $k$-th category. Using this approach, we are able to qualify top $K$ users in each category by sorting the interest score $u_{i, cat=k}^t$. Depending on the advertiser's goals and the category, choice of $K$ ranges from campaign to campaign. We note that a single user can be qualified into one or more interest categories (e.g., user can be categorized in ``Sport", ''Sport/Basketball", and ''Health and Nutrition/Vitamins" at the same time), and, when the system was deployed, a user was assigned to $13$ categories on average. An example of user profiles qualified into certain categories is given in Table \ref{tab:infres}.

\subsubsection{Leveraging the follower graph}

To be able to target Tumblr users who do not create much content, but actively follow and engage with other blogs, we leverage the follower graph to create additional categorized features. Using equation \ref{intensity_cat} we can identify frequent bloggers in certain interest categories by focusing on a small percentage of users with maximum $u_{i, cat=k}$. Following and liking posts created by social influencers in the $k$-th category serves as additional evidence of one's interest in that category.

In each interest category, we label the $5\%$ of users with the highest number of activities in that category, as frequent bloggers. Next, we update the interest score of all users $u_i$ in the $k$-th category, in the following manner.
\begin{equation}\label{intensity_cat_add}
u_{i, cat=k}^t += \sum_{b \in \mathcal{F}_i} \sum_{eng \in b} \alpha^{(t-t_{eng})} ~w_{eng} ~I(f \text{ is of class }k),
\end{equation}
where $\mathcal{F}_i$ is a set of all frequent blogs followed by user $u_i$, $eng$ are all engagements with $b$-th blog, i.e. likes or follow actions, along with their weights $w_{eng}$ (e.g., if the posts created by $b$-th blog was liked ten times, then the value is equal to 10; if user followed the $b$-th blog, then the value is equal to 1), and indicator function $I(\cdot)$ returns 1 is the blog is of class $k$, and 0 otherwise. Similarly to other activities, we applied the exponential decay to the sum, based on the timestamps of follow and like actions.

We have observed that the additional signals, in form of follow and like engagement with frequent bloggers, increases our segment sizes, making it possible to efficiently target more users.
  
\subsection{Results}
To evaluate the generated interest profiles we performed online A/B testing, and worked with several advertisers who ran concurrent interest-targeted and untargeted campaigns. We tracked user engagement with their ads in terms of sponsored post likes, reblogs, and follows, and show the results for $8$ targeting campaign in Table \ref{tab:ab_test_resultsl}. We observed an average lift of $20\%$ in user engagement (aggregate of $3$ metrics) with sponsored posts as compared to untargeted campaigns. This performance result represents a significant improvement over the baseline approach.

\section{Gender Prediction}
In this section we explain details of our gender prediction model, based on the user profiles described in previous sections. We first describe the generation process of a golden set of labeled users, used to train a predictive model that generalizes well on the remaining unlabeled users. This is followed by the model description and discussion of results.

\subsection{Collecting ground-truth labels}
In order to train machine learning method for gender prediction, in addition to user profiles we also require labels that present the ground truth (i.e., ``male'' or ``female''). However, Tumblr does not collect gender information when users sign-up, leaving open a question on how to obtain such data.  

\begin{table}[t]
\centering
{
\caption{A/B test results on $10\%$ of user population}
\begin{tabular}{ lcc }
{\bf Campaign} & {\bf Control} & {\bf Targeted}\\
\hline
\rowcolor{lightgray}
Home and Garden & $-$  & $+9.71\%$ \\ 
Style \& Fashion & $-$ &  $+42.53\%$ \\
\rowcolor{lightgray}
Sports/Outdoor Sports & $-$  & $+19.86\%$ \\ 
Arts \& Enter./Television & $-$ &  $+24.37\%$ \\
\rowcolor{lightgray}
Arts \& Enter./Video Games & $-$  & $+19.02\%$ \\ 
Pets/Dogs & $-$ &  $+27.21\%$ \\
\rowcolor{lightgray}
Arts \& Entertainment 1 & $-$  & $+9.08\%$ \\ 
Arts \& Entertainment 2 & $-$ &  $+6.54\%$ \\
\bottomrule
\end{tabular}
\label{tab:ab_test_resultsl}
}
\end{table}

To address this problem, we proposed to leverage highly informative blog description data to infer user gender information. In particular, very often users declare their name in the blog description, as illustrated in Figure \ref{fig:blog_titles}. To extract user declared names, we used several regular expression rules that we found to result in very high precision. The obtained results from a large set of name-matching regular expressions were editorially tested for quality. tt was found that regular expressions reported in Table \ref{tab:desc_keywords} yielded the most reliable extracted names (valid names were extracted in more than $95\%$ of cases).
Then, to generate the gender ground truth, we used the US census data of popular baby names\footnote{\url{www.ssa.gov/oact/babynames/limits.html}} from year 1880 to 2013 to created a ``name $\rightarrow$ gender'' mapping. As some names are given to both males and females, we used the empirical counts of babies with certain name to generate labels. More specifically, we used male/female empirical ratios as soft labels, with $1$ indicating 100\% confidence in male and $0$ indicating 100\% in female name.  

\begin{table}[t]
\centering
{\footnotesize
\caption{Matching names in blog description (* represents matched name; following methodology described in text, $564{,}000$ female and $395{,}000$ male users were found) }
\begin{tabular}{ l c c }
{\bf regex} & {\bf count} \\
\hline
\rowcolor{lightgray}
my name is * & 783,564 \\
my name's * & 291,811 \\
\rowcolor{lightgray}
me llamo * & 47,663 \\
the name's * & 38,065 \\
\rowcolor{lightgray}
mi nombre es * & 9,751 \\
mi chiamo * & 9,181 \\
\rowcolor{lightgray}
mein name ist * & 1,025 \\
meu nome e * & 512 \\
\rowcolor{lightgray}
mon nom est * & 215 \\
mio nome e * & 185 \\
\bottomrule
\end{tabular}
\label{tab:desc_keywords}
}
\end{table}

\subsection{Proposed approach}
Let ${\cal D}_g = \{ ({\bf x}_i, y_i), i = 1, ..., N \}$ denote our gender data set, where $N$ is the total number of labeled users, ${\bf x}_i$ is a $K$-dimensional user feature vector generated from the user profiles, and $y_{i} \in [0, 1]$ is the user label (real-valued number, ranging from $0$ to $1$). The feature vectors were generated from user profiles described in Section \ref{sec:kw_ext}, by disregarding time stamps (due to the fact that, unlike users' interest, their gender does not fluctuate), and using the keywords as features and overall keyword counts as feature values. Our goal is to learn a gender-predictive model, $f: {\bf x} \rightarrow y$. 
As a classification model we used logistic regression, parameterized by a weight vector ${\bf w}$. We assume that the posterior gender probabilities can be estimated as a linear function of input ${\bf x}$, passed through a sigmoidal function,
\begin{equation}\label{posterior}
\P(y=1|{\bf x}) = f({\bf x}, {\bf w}) = \frac{1}{1+\exp(-{\bf x}^\T {\bf w})},
\end{equation}
and $\P(y=0|{\bf x}) =1-\P(y=1|{\bf x})$. To estimate the parameters ${\bf w}$, we minimize the following loss function,
\begin{equation}\label{weighted_mse}
\min_{ {\bf w} \in \R^K} \frac{1}{N}\sum_{i=1}^{N} \big(y_i - f({\bf x}_i, {\bf w})\big)^2 + \lambda \| {\bf w} \|_1,
\end{equation} 
where hyper-parameter $\lambda$ controls the $\ell_1$-regularization, introduced to induce sparsity in the parameter vector and reduce the feature space to a subset of features that are the most predictive. 
For data sets with a large number of features, as in our use case, it is common that many features are not useful for producing a desired learning result. For this reason, the $\ell_1$-regularization was critical part of our training procedure. In addition, we experimentally observed that the model generalizes better when we trained an initial model with $\ell_1$-regularization to find which features have non-zero weight, and then do another round of training without $\ell_1$-regularization by only using features with non-zero weights from the first round to learn a better classifier.

Given a trained logistic regression model, the posterior class probabilities are estimated as $f({\bf x}_i, {\bf w}) \in \lbrack 0, 1 \rbrack$. Then, the classification predictions are made by thresholding, as $\hat{y}_i = \text{sign}(f({\bf x}_i, {\bf w}) - \theta)$, where threshold $\theta$ is set between $0$ and $1$ to ensure desired precision and recall according to specific advertisers requirements.

\subsection{Results}

\begin{table}[t]
\centering
{
\caption{Accuracy of gender model on hold-out set}
\begin{tabular}{ ccccc }
{\bf Gender} & {\bf Threshold} & {\bf Precision} & {\bf Recall}\\
\hline
\rowcolor{lightgray}
female & $<0.5$ & 0.806 & 0.838 \\ 
male & $>0.5$ & 0.794 & 0.689 \\
\bottomrule
\end{tabular}
\label{tab:main_res}
}
\end{table}

\begin{table}[t]
\centering
{
\caption{Editorial evaluation of random user predictions}
\begin{tabular}{ cccc }
{\bf Class Prediction} & {\bf Correct} & {\bf Wrong} & {\bf Not sure}\\
\hline
\rowcolor{lightgray}
female & $429$ & 4 & 298 \\ 
male & $144$ & 5 & 127 \\
\bottomrule
\end{tabular}
\label{tab:main_res_editorial}
}
\end{table}

To evaluate the accuracy of our gender prediction framework, we trained a logistic regression model on $70\%$ of golden set and tested on the remaining $30\%$. We used Vowpal Wabbit \cite{langford2009sparse} implementation on Hadoop to train the model. To illustrate the performance of our gender classifer, performance results in terms of precision and recall measures for threshold $\theta=0.5$ are presented in Table~\ref{tab:main_res}. 

In addition to evaluation on the hold-out set, we also editorially evaluated gender predictions on the unlabeled data set of user profiles. We randomly picked $1{,}007$ gender predictions from the population of $64.1$ million users and asked editors to visit their profiles and verify the gender. They were instructed to mark our predictions as ``correct", ``incorrect", or ``not sure". The ``not sure" grade is to be used when visual inspection of profile is not conclusive, as we found was often the case. The editorial judgment came back with $573$ ``correct" ($429$ females and $144$ males), $9$ ``incorrect", and $425$ ``not sure" grades (see Table~\ref{tab:main_res_editorial}). The fact that there are so many ``not sure" grades indicates that in many cases it is hard to infer the gender even after manual efforts, further indicating the benefits of the proposed approach and its superior performance when compared to humans. Finally, we retrained the models with $100\%$ of golden set and deployed it in Yahoo production systems. 

{\bf Demonstration video} of the most predictive tags in each gender group is available online at \url{https://www.youtube.com/watch?v=jXGJ0TpOlhg}.

\section{Deployed system}
Due to a rapid growth of Tumblr and large number of activities generated by the existing users, we implemented daily scoring of users in Yahoo production servers. We store the activities in Hive tables\footnote{https://hive.apache.org} for efficient retrieval. The decayed counts used in interest prediction are updated on a daily basis by multiplying the old feature values by the decay factor $\alpha$ and adding new activities. To infer gender of new users we implemented daily scoring by leveraging MapReduce on Hadoop\footnote{https://hadoop.apache.org}. Both interest and gender models are retrained on a regular basis.

After thorough editorial evaluation of the inferred gender and interest targeting, both targeting frameworks were enabeled through Gemini self-serve tool\footnote{https://gemini.yahoo.com}. Advertisers can choose to use gender and/or interest targeting with custom segment sizes, allowing for effective targeting campaigns.

\section{Conclusions}
We presented the steps in developing a large-scale Tumblr gender and interest targeting framework, where we used historical Tumblr activities to create rich user profiles. We described the methodology, including a novel semi-supervised neural language model, and the high-level implementation details behind the deployed system. Currently, our gender and interest predictions cover more than $90$\% of Tumblr daily activities, and are heavily leveraged by advertisers. In our ongoing work, we are concentrating on creating custom keyword-targeted advertising segments, specifically tailored for a particular advertiser, including addressing problems of keyword discovery and expansion.

\balance

\bibliographystyle{abbrv}
\bibliography{tumblr_refs}
\end{document}